\documentclass[journal]{IEEEtran}

\usepackage{booktabs} 

\usepackage{amsfonts}
\usepackage{epstopdf}
\usepackage{sidecap}
\usepackage{wrapfig}
\usepackage{graphicx}
\usepackage{subfigure}
\usepackage{color}
\usepackage{wrapfig}
\usepackage{caption}
\usepackage{paralist}
\usepackage{amssymb}
\usepackage{adjustbox}

\usepackage{amsmath,epsfig}
\usepackage{color}
\usepackage{algpseudocode} 
\usepackage{algorithm}
\usepackage{enumitem}
\begin{document}
%
\title{Haptic-enabled Mixed Reality System for 
	Mixed-initiative Remote Robot Control}
%
%
%

\author{Yuan~Tian,
        Lianjun~Li,
        Andrea~Fumagalli,~\IEEEmembership{Member,~IEEE,}
        Yonas~Tanesse,~\IEEEmembership{Member,~IEEE,}
        and Balakrishnan~Prabhakaran,~\IEEEmembership{Member,~IEEE,}
\thanks{Yuan Tian is with OPPO Research Center, Palo Alto, CA 94303, USA email: {yuan.tian@oppo.com}}
\thanks{Lianjun Li, Dr. Andrea Fumagalli, Dr. Yonas Tanesse, Dr. Balakrishnan Prabhakaran are with the University of Texas at Dallas, Richardson, TX 75080. email: {lianjun.li1, andreaf, yonas.tadesse, bprabhakaran}@utdallas.edu}
}
\maketitle

\begin{abstract}
Robots assist in many areas that are considered unsafe for humans to operate. For instance, in handling pandemic diseases such as the recent Covid-19 outbreak and other outbreaks like Ebola, robots can assist in reaching areas dangerous for humans and do simple tasks such as pick up the correct medicine (among a set of bottles prescribed) and deliver to patients. In such cases, it might not be good to rely on the fully autonomous operation of robots. Since many mobile robots are fully functional with low-level tasks such as grabbing and moving, we consider the mixed-initiative control where the user can guide the robot remotely to finish such tasks. 
In this paper, we proposed a novel haptic-enabled mixed reality system, that provides haptic interfaces to interact with the virtualized environment and give remote guidance for mobile robots towards high-level tasks. The system testbed includes the local site with a mobile robot equipped with RGBD sensor and a remote site with a user operating a haptic device. A 3D virtualized real-world static scene is generated using real-time dense mapping. The user can use a haptic device to “touch" the scene, mark the scene, add virtual fixtures, and perform physics simulation. The experimental results show the effectiveness and flexibility of the proposed haptic-enabled mixed reality system.
\end{abstract}


%
\IEEEpeerreviewmaketitle

\section{Introduction}
\begin{figure}[htb!]
	\centering
	\includegraphics[width = 1.0\linewidth]{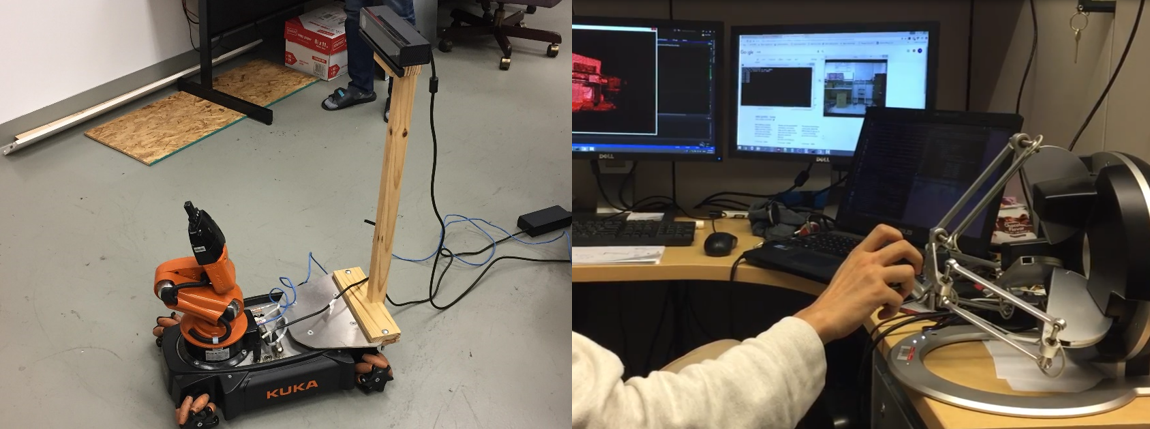}
	\caption{The system setup in our proposed work. Left is local site: KUKA youBot with a Kinect V2 placed on top. Right is the remote site, the user interacted with the 3D reconstructed scene and guide the robot using haptic device. }
	\label{fig:hapticrobot}
	\vspace{-1em}
\end{figure}

Networked mixed reality has become popular for all kinds of the applications such as distributed collaborations~\cite{rios2016exploring}, training~\cite{gonzalez2017immersive}, and video streaming~\cite{han2011mixed}. Such a networked mixed reality system can merge the real and virtual worlds to produce new environments where physical and virtual objects interact with each other in real-time.
Many researchers have applied mixed reality for robot teleoperations~\cite{yang2016teleoperation,li2017haptic}, human-robot interaction control~\cite{robert2011exploring, wang2011mixed,kebria2019robust} and mixed-initiative control~\cite{sauer2011mixed, cacace2014mixed}. Among all these robot controls, mixed-initiative control is a hot topic that has drawn much attention. \cite{heger2006sliding, sellner2006coordinated, carlone2010steps} introduced the different levels of autonomy: full autonomy (robot-initiative), mixed-initiative and teleoperation (human-initiative).
In reality, more control systems are designed with the capability to ``sliding autonomy", which means the system supports the seamless transfer to different levels of control. Mixed-initiative control is to give the robot high-level commands instead of teleoperation, as shown in Figure~\ref{fig:hapticrobot}. It becomes more important in particular for improving situational awareness,
decreasing the workload of the human operator and at the same time guaranteeing safe operations. 


Some previous methods have introduced haptic feedback and mixed reality for mixed-initiative control~\cite{sauer2011mixed, cacace2014mixed}, most of them applied haptic device as a multidimensional teleoperation controller or used haptic guidance force and environment force as the feedback of robot motions. 
In this paper, we consider how 3D haptic interaction in the mixed reality could be expanded to help the robot mixed-initiative control. 
Haptic interaction with the 3D virtual environment is very popular in computer graphics applications to provide immersive experiences, which is to use a haptic avatar to explore freely in a 3D world and interact with 3D objects (push, touch) and feel the force feedback.
Combining haptic interaction with mixed-initiative control will provide more flexibility towards the control.

Several high-quality sensors can be used to provide the mapping and localization information for mobile robots, such as Light Detection and Ranging (LIDAR), RGBD cameras, etc. The sensor data can generate the virtualized real-world scene with dense geometry~\cite{kelly2011real}. Users can use a mouse cursor, joystick, or any other input device to operate the virtualized objects in a mixed reality environment to realize some goals~\cite{le2006distributed}. The introduction of the haptic interaction into the mixed reality environment will add more flexibility to operations. Haptic devices provide more degrees of freedom for cursor motions and provide force feedback for more immersive experiences. Using haptic devices, users can remotely ``touch'' and mark the virtualized environment from the streaming data~\cite{tian2017real}. Furthermore, the haptic interface can be integrated with the physics simulation of objects, the virtualized object can be moved in the scene. These operations can provide more flexible control and guidance for the mobile robot since the robot also uses dense mapping for localization and navigation~\cite{ganganath2012mobile}. 

In this paper, we assume the robots have the full functionality for low-level tasks such as grabbing, moving based on the input 3D objects and positions. To have such a haptic-enabled mixed reality system for mixed-initiative remote control, there are several challenges:
\begin{enumerate}[leftmargin=*, label=(\roman*)]
	\item The first challenge is real-time streaming of virtualized data over a network that is susceptible to data loss and delays. 
	\item The second challenge comes from the requirement of the haptic rendering with the 3D virtualized environment, which needs to be robust, efficient, and smooth. 
	\item Object segmentation from the 3D scene that will be needed. The segmented object will be the input for robot grabbing. 
	\item Furthermore, network latency will delay the guidance commands from the server, and hurt the haptic interactions, which might lead to the disparity of goal motions and real-world motions for the robot.
\end{enumerate}

To address these challenges, we have proposed a novel haptic-enabled mixed reality system for mixed-initiative remote control. The system provides a haptic interface to interact with the virtualized environment and gives remote guidance for mobile robots towards high-level tasks. The system includes a local site with a mobile robot equipped with an RGBD sensor and a remote site with a user operating a haptic device. A 3D virtualized real-world static scene is generated using real-time dense mapping. The user can use a haptic device to remotely ``touch" the scene, mark the scene, add virtual fixtures, and perform physics simulation. Specifically, the technical contributions of our method are as follows:
\begin{itemize}[leftmargin=*]
	\item A real-time efficient and robust mixed reality platform for mixed-initiative control is proposed to enable haptic interactions with streaming data.   
	\item A TSDF-based haptic rendering method with the streaming surface is proposed to ensure the smooth and robust haptic interaction with a virtualized static scene. 
	\item A superpixel-enhanced instance segmentation method is proposed to segment objects fast and accurately.
	\item Different types of haptic interfaces are introduced in the mixed reality platform, and a robot state prediction method is proposed to compensate network delays. 
\end{itemize}

\section{Related Work}
\label{sec:relatedwork}

Researchers have worked hard towards the haptic-enabled Tele-operation field to deal with the challenges of communication delay, control strategy~\cite{yang2016teleoperation,li2017haptic,kebria2019robust}.  Networked mixed reality platform is widely introduced to provide a remote immersive platform to improve the robot control, and enable the interactions between the physical robots and virtual objects~\cite{ chouiten2012distributed, kelly2011real, wu2020mixed}. 
\cite{robert2011exploring} presented a mixed reality (MR) platform: an integrated physical and virtual environment, in which the user interacts with a teleoperated robot by passing a graphical object. 
\cite{wang2011mixed} proposed a mixed reality interface for a remote robot
using both real and virtual data acquired by a mobile equipped with an
omnidirectional camera and a laser scanner. The MR interface can enhance the
current remote robot teleoperation visual interface.
Son et al. \cite{son2013human} investigated 3 haptic cues for bilateral teleoperation of multiple mobile robots, and found force cue feedback best support the maneuverability. 
In~\cite{kelly2011real}, dense geometry and appearance data were used to generate a photorealistic synthetic exterior line-of-sight view of the robot including the context of its surrounding terrain. This technique converted remote teleoperation into a line-of-sight remote control with the capacity to remove latency. Chouiten et al.~\cite{chouiten2012distributed} proposed a distributed mixed reality system that implemented the real-time display of digital video stream to web users, by mixing 3D entities with 2D live videos by a teleoperated ROV.  
Some methods introduced the haptic feedback into the mixed reality control platforms.
A mixed reality system~\cite{sauer2011mixed} is developed with GUI interfaces and force feedback including path guidance forces, collision preventing forces or environmental force to improve the performance of high-level tasks operations. Later, Cacace et al.~\cite{cacace2014mixed} proposed a mixed-initiative control system to add a human loop to control the velocity of aerial service vehicles, and force feedback is used to enhance the control experience. Different from these methods, our system introduced the haptic interaction with 3D scenes into mixed reality, which will bring more flexibility. Lee et al.~\cite{lee2017visual} proposed a visual guidance algorithm that dynamically manipulates the virtual scene to compensate for the spatial discrepancies in a haptic augmented virtuality system. In \cite{wu2020mixed}, an interface based on Microsoft HoloLens is proposed, which can display the map, path, control command, and other information related to the remote mobile robot, also provide interactive ways for robot control.

Recently, many robots are equipped the depth sensors for localization and mapping~\cite{cunha2011using, ganganath2012mobile, henry2012rgb}.
KinectFusion\cite{izadi2011kinectfusion,oleynikova2017voxblox} is one of the most popular methods, which fuses the streaming RGBD data from the Kinect camera, and saves it as a Truncated Signed Distance Function (TSDF).      
KinectFusion can provide the full-scene dense geometry to enable mixed reality. 
\cite{tian2017real} firstly introduced the real-time haptic rendering with streaming deformable surface generated by KinectFusion. Besides the simulation of surface deformation, this method provided a haptic rendering pipeline including collision detection, proxy update, and force computation.
The collision detection is performed by ray casting in the TSDF data structure that is saved in GPU. At each time step, based on the haptic interaction position (HIP), this method finds the corresponding proxy: the nearest surface point to HIP. Finally, the haptic force is computed by the positions of proxy and HIP. This method is computationally efficient and integrated well with the KinectFusion framework. However the force rendering of this method only works well with the planes, and it will be unstable at the intersecting boundary of two or more planes. In the real-world scenes, the complex geometry nature harnesses the stability using this method. We borrowed the idea of haptic interaction with the streaming dense surface and proposed a new pipeline of haptic rendering to keep both the stability and efficiency.

Object segmentation from images has been an essential topic for scene understanding computer vision society~\cite{adams1994seeded, deng2001unsupervised}. Later, RGBD images are used for semantic mapping: which includes both dense mapping (like Kinect fusion) and object detection, semantic classification
~\cite{silberman2012indoor, hermans2014dense, muller2014learning, gupta2015indoor, he2017std2p}. 
Our method aims to interact with the reconstructed object surface, therefore we chose to only segment the object in real-time, rather than semantic classification.
The segmentation algorithms include many categories: region growing~\cite{adams1994seeded, deng2001unsupervised}, clustering~\cite{achanta2010slic, silberman2012indoor}, and deep learning based methods~\cite{gupta2014learning,valentin2015semanticpaint, he2017std2p}.
In our system, since we used haptic interaction, therefore we developed an interactive region growing method for object segmentation using both color image and depth image, and fuse the information into the TSDF data structure.

\section{Overview}
\label{sec:overview}

\begin{figure}[htb!]
	\centering
	\includegraphics[width = 1.0\linewidth]{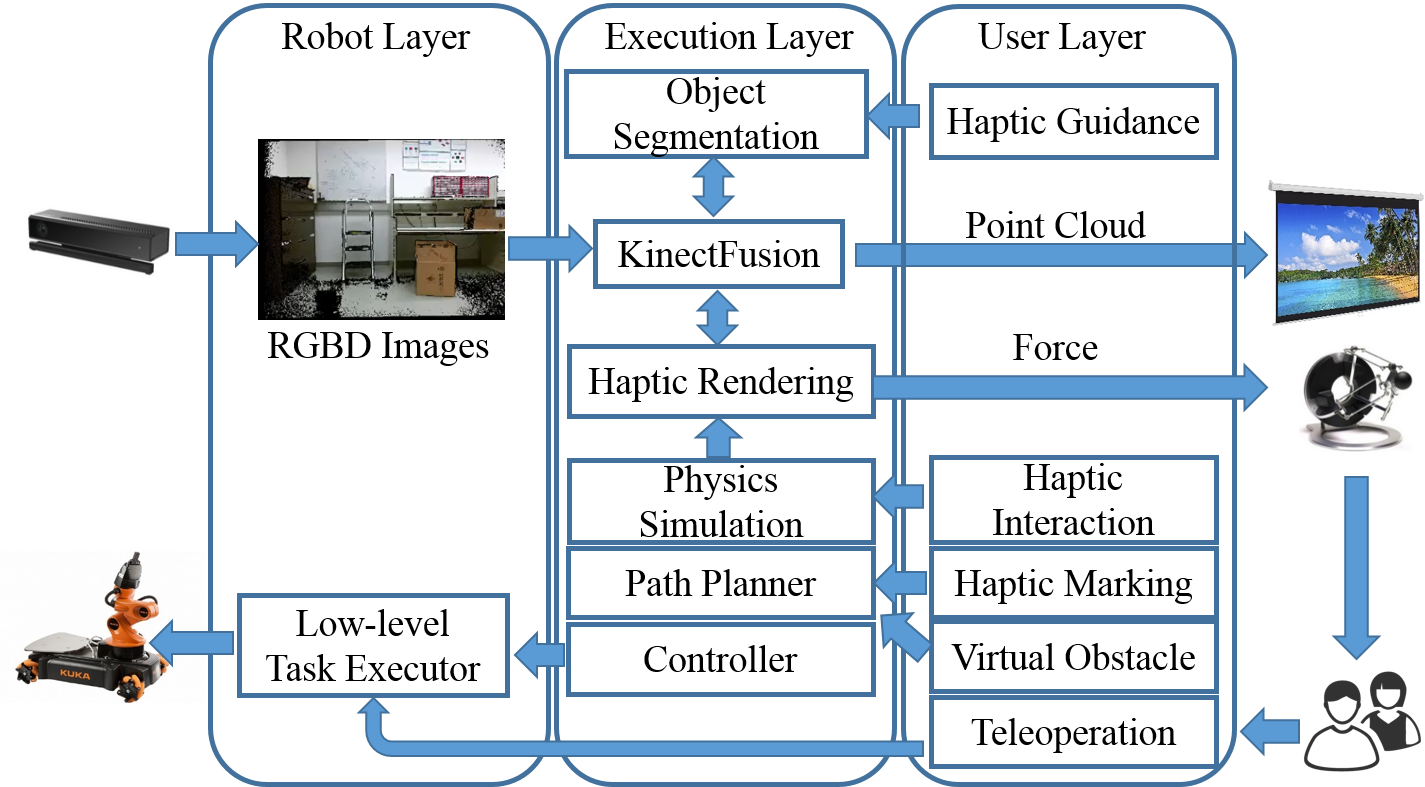}
	\caption{The proposed system architecture.  }
	\label{fig:piplineRobot}
	\vspace{-1em}
\end{figure}

As shown in the Figure~\ref{fig:piplineRobot}, our mixed reality system comprises three layers: 
\begin{itemize}[leftmargin=*]
\item The Robot Layer connected with a mobile robot (in our implementation, it is KUKA youBot) and a Kinect V2 placed on the top of the robot. This layer collects the color and depth images and sends them to the Execution Layer. There is a low-level task executor to execute the control commands that are sent by the controller in the Execution Layer. 
\item The Execution Layer gets the RGBD images and performs simultaneous localization and mapping (SLAM) using KinectFusion~\cite{izadi2011kinectfusion}. Then KinectFusion generates a point cloud every time step for visual rendering. It will combine the object segmentation module to segment and mark the object if necessary. The layer also includes a separate thread for haptic rendering. This module will compute the force feedback and send it to a haptic device. The physics simulation module handles the situation that a haptic interaction interface is enabled to interact with a virtual object. The Execution Layer also includes the path planner to generate a path based on the user's marking or the virtual obstacles that the user adds. The controller module is used to generate a predicted position and commands for the robot to follow. 
\item The User Layer provides all the interfaces and outputs. The user can either use the teleoperation interface to directly operate the robot, or use haptic interfaces to interact with the 3D environments. Haptic guided object segmentation interface is only used for segmentation. Haptic interaction interface enables the user to use haptic to push the virtual object as the target or the obstacle. Haptic marking can either define a path on the ground or mark one object, then the robot will try to follow the path or approach the object. The virtual obstacle interface enables the user to add virtual obstacles (any form of geometry) into the scene, then the path planner will search for a new path to avoid the obstacle.	
\end{itemize}

The Robot Layer and Execution Layer are connected over the Internet. The dense mapping is done in Execution Layer instead of Robot Layer since the RGBD image data has a smaller size than the 3D point cloud. Another reason is that haptic rendering with TSDF data has very good performance \cite{tian2017real}.

In the real world, the system includes two sites: local site and remote site, that are connected with high speed Internet (10 Mbps), TCP/IP protocol is applied for the data transfer. A Kinect v2 is placed onto the KUKA youBot, and it is connected with a Linux machine to where the Robot Layer belongs to. The system transfers the RGBD images to the remote site by 15-20 fps. At the remote site, the server machine implements the Execution Layer and User Layer.


\section{TSDF-based Haptic Rendering with 3D Streaming Environment}
\label{sec:hapticrendering}

\begin{figure}[htb!]
	\centering
	\includegraphics[width = 1.0\linewidth, height = 5em]{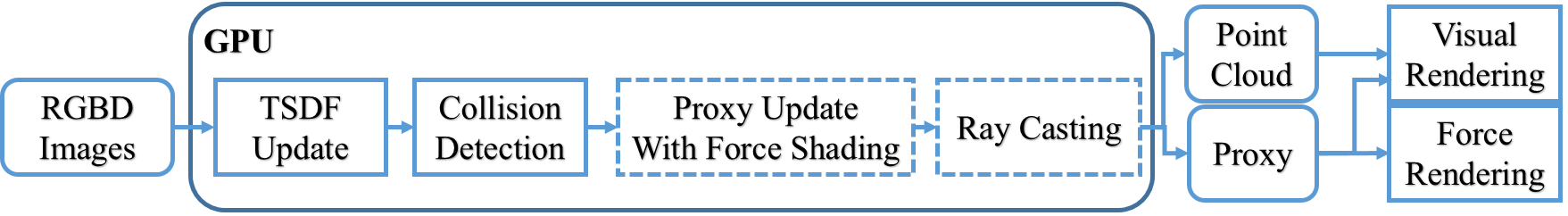}
	\caption{The haptic rendering pipeline: collision detection method~\cite{tian2017real} is used; a new proxy update method is used to find proxy; Friction and texture are added to simulate the properties.}
	\label{fig:hapticrendering}
	\vspace{-1em}
\end{figure}

Since KinectFusion is applied for dense mapping and localization, dense geometry is generated as the streaming surface of the 3D virtualized environment. 
Inspired from the work \cite{tian2017real} of haptic rendering, we proposed a new processing pipeline to handle the proxy update, as shown in Figure~\ref{fig:hapticrendering}. We also proposed a novel proxy update method with force shading, which is more efficient and guarantees stable rendering in the intersecting boundaries of different planes. Furthermore, we proposed the method to add surface properties such as friction and haptic textures. 

\subsection{Proxy Update with Force Shading}
The proxy update is the most important part of the constraint-based haptic rendering~\cite{ruspini1997haptic}, since the proxy is not only used to compute the force but also rendered visually to the viewers. If the proxy update is not stable and smooth, the force rendering and visual rendering will not be smooth.

\begin{figure}[htb!]
	\centering
	\includegraphics[width = 1.0\linewidth]{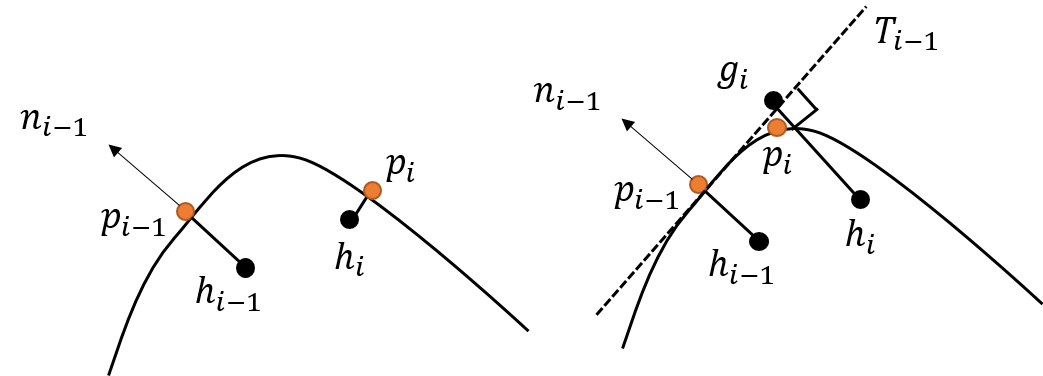}
	\caption{Left is the proxy update method proposed in ~\cite{tian2017real}; right is our proxy update method using force shading. For the same movement, the left has a jump change of proxy positions, and the right has much smoother proxy updates. }
	\label{fig:forceshading}
	\vspace{-1em}
\end{figure}
In \cite{tian2017real}, the proxy update uses a gradient-based method to find the nearest surface point. As shown in Figure~\ref{fig:forceshading}, the left figure shows a scenario that haptic interact with a surface with a sharp change, which is like the intersecting boundary with two flat planes. In this scenario, the haptic interaction point (HIP) is moved by the user from $h_{i-1}$ to $h_i$, the proxy position is changed from $p_{i-1}$ to $p_i$. Since the proxy is always the nearest surface point according to HIP, the proxy has a sudden change in position. It would feel as though it ``jumps" to the other side of the surface, and computed force is changed vigorously to an almost reversed direction.

In this paper, we proposed a proxy update method with force shading. Force shading was first introduced into haptic rendering in~\cite{ruspini1997haptic}, which borrows the idea from Phong shading rendering in computer graphics applications. Our method will handle two scenarios: 

a. If the HIP is the first contact to the surface, the proxy is to find the nearest surface point. Instead of the gradient-based iterative method proposed in~\cite{tian2017real}, we integrate the task of finding the nearest surface point into the ray casting step in KinectFusion. The reason is that our application will not consider the deformable property for the surface, therefore the ray casting is performed after the haptic rendering. Per-pixel ray marches in TSDF to generate the point cloud for the whole surface. During this procedure, the distances between the HIP and every point on the surface are computed and saved. The nearest surface point finding problem now becomes a parallel problem that finds the minimum in the distance array. This problem can be solved through parallel reduction~\cite{harris2007optimizing}. This algorithm is shown in Algorithm.~\ref{alg:nearest}.

\begin{algorithm}
	\caption{Nearest Surface Point Finding in Ray Casting}
	\label{alg:nearest}
	\begin{algorithmic}[1]
		\Require {Given the starting point $h$}
		\State{Parellelized thread: each pixel's corresponding ray}
		\State {Marches from minimum depth, stop when zero crossing to get surface point $s$}
		\State {Compute the distance $d = |s - h|$}  
		\State {Parallel reduction to get minimal distance, return the corresponding surface point as the nearest one}
	\end{algorithmic}
\end{algorithm}

b. After the HIP penetrates into the surface, the subsequent proxy position needs to be updated since the HIP will penetrate further into the volume. As shown in Figure~\ref{fig:forceshading}, the nearest surface point is not appropriate for this scenario, a more correct way is to constrain the successive proxy. The previous time step normal $n_{i-1}$ is used to define a tangent plane, the normal of proxy will be computed every time step. Tracking this normal is like tracking a tangent gliding plane over the surface physically. As shown in the right of Figure~\ref{fig:forceshading}, the tangent plane $T_{i-1}$ is ``dragged" by the new proxy position $h_i$ while attached on the surface. So, the tangent plane can be treated as a constraint plane for the proxy. First, we drop a perpendicular from $h_i$ to this constraint plane to get a goal position $g_i$, which is the first approximation of the proxy. Then, the nearest surface finding in the ray casting step will find the new proxy $p_i$. This two-step method is similar to the force shading method~\cite{ruspini1997haptic}. The core of the method is to use the tangent plane to constraint the new proxy in a physically plausible way, then refine it as the nearest surface point. The whole procedure is shown as Algorithm.~\ref{alg:proxyupdate}.  

\begin{algorithm}
	\caption{Subsequent Proxy Update with Force Shading}
	\label{alg:proxyupdate}
	\begin{algorithmic}[1]
		\State {Based on the normal $n_{i-1}$, get the tangent plane $T_{i-1}$}
		\State {Drop a perpendicular from current HIP $h_i$ to $T_{i-1}$ to get $g_i$}
		\State {Use Algorithm.~\ref{alg:nearest}  initialized from $g_i$ to compute the final proxy $p_i$}
	\end{algorithmic}
\end{algorithm}
\vspace{-2em}

\subsection{Surface Properties}
\begin{figure}[htb!]
	\centering
	\includegraphics[width = 1.0\linewidth]{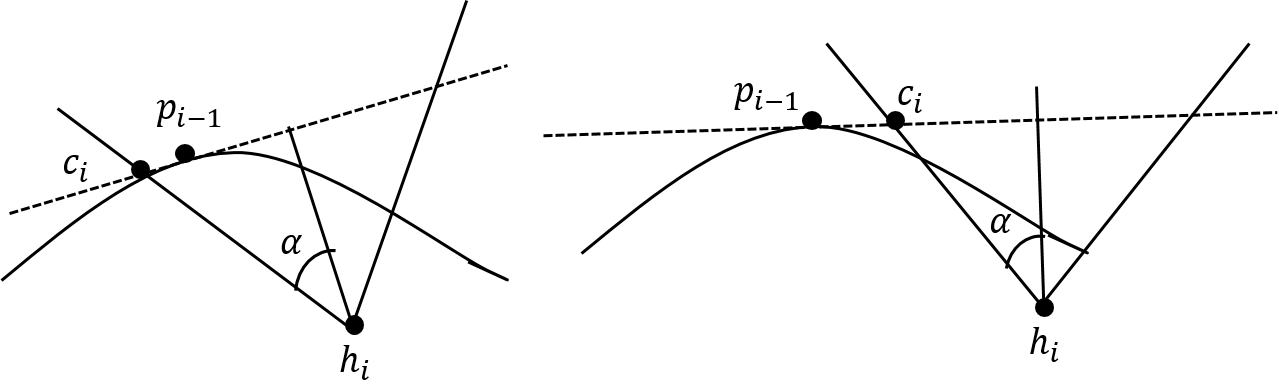}
	\caption{Proxy update for friction. Left simulates the stick force: when the previous proxy is inside the friction cone, the proxy will not be updated; right simulates slip force: when the previous proxy is outside, the proxy will be updated as the nearest boundary point on the boundary cone.}
	\label{fig:friction}
	\vspace{-2em}
\end{figure}

Surface properties can simulate friction force and differential haptic textured surface. 
Similar to~\cite{salisbury1997haptic}, the friction force can be simulated by a simple change using the known friction cone. The angle $\alpha$ defines a cone starting from the current HIP $h_i$, as shown in Figure\ref{fig:friction}. The friction cone has an interaction circle with the tangent plane. $\alpha = arctan(\mu)$, where $\mu$ is a user-defined friction coefficient. If the previous time step proxy $p_{i-1}$ is inside the circle, the new proxy will be directly set the same as before: $p_i = p_{i-1}$. If outside, then the goal position (approximated proxy) $g_i = c_i$, where $c_i$ is the point closest to $p_{i-1}$ on the circle. The two scenarios correspond to static friction and dynamic friction. The haptic texture can be easily extended by using the bump texture method~\cite{blinn1978simulation}. It can generate constraints for each point to change the normal. 

\section{Interactive 3D Object Segmentation}
\label{sec:objectSegmentation}

It is very necessary to provide the interface to segment 3D objects in the scene. Such an interface enables more flexible haptic interaction, e.g. haptic texture, material properties for different objects, and also provides the object position and orientation for robot grasping tasks. Many researchers combine object detection and semantic classification into dense mapping
~\cite{silberman2012indoor, hermans2014dense, muller2014learning, valentin2015semanticpaint, gupta2015indoor, he2017std2p}. Our system aims to build haptic-enabled interfaces for the mixed-initiative control, therefore the high-level semantic segmentation is beyond our scope. We propose an interactive 3D object segmentation method that is not only efficient but also compatible with the popular high-level object semantic algorithms as the input. 

\begin{figure}[htb!]
	\centering
	\includegraphics[width = 1.0\linewidth]{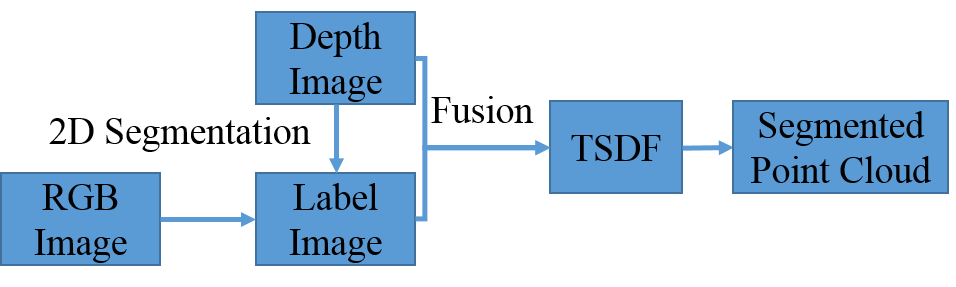}
	\caption{ The proposed method for interactive 3D object segmentation. }
	\label{fig:segpipeline}
	\vspace{-1em}
\end{figure}

The straightforward way is to segment the 3D object from the 3D point cloud. It is also possible to use the KD-tree to speed up the neighbor search for points. This method takes extra processing time. Another way is to perform the segmentation based on TSDF data and save the segmentation information into the TSDF. In the KinectFusion pipeline, the depth image is fused for surface reconstruction at each time step. Based on this observation, we propose a two-phase algorithm. In the first phase, the 2D segmentation is performed from both depth image and color image. After the 2D segmentation, a label image $L_i$ is generated. In the second phase, the segmentation is fused into the TSDF together with the depth image. In this way, the segmentation is seamlessly integrated into the KinectFusion and reduces the time cost. Moreover, the segmentation information will be fused by weight, which generates robust segmentation results. The whole pipeline is shown as Figure~\ref{fig:segpipeline}.

\begin{figure}[htb!]
	\centering
	\includegraphics[width = 0.8\linewidth]{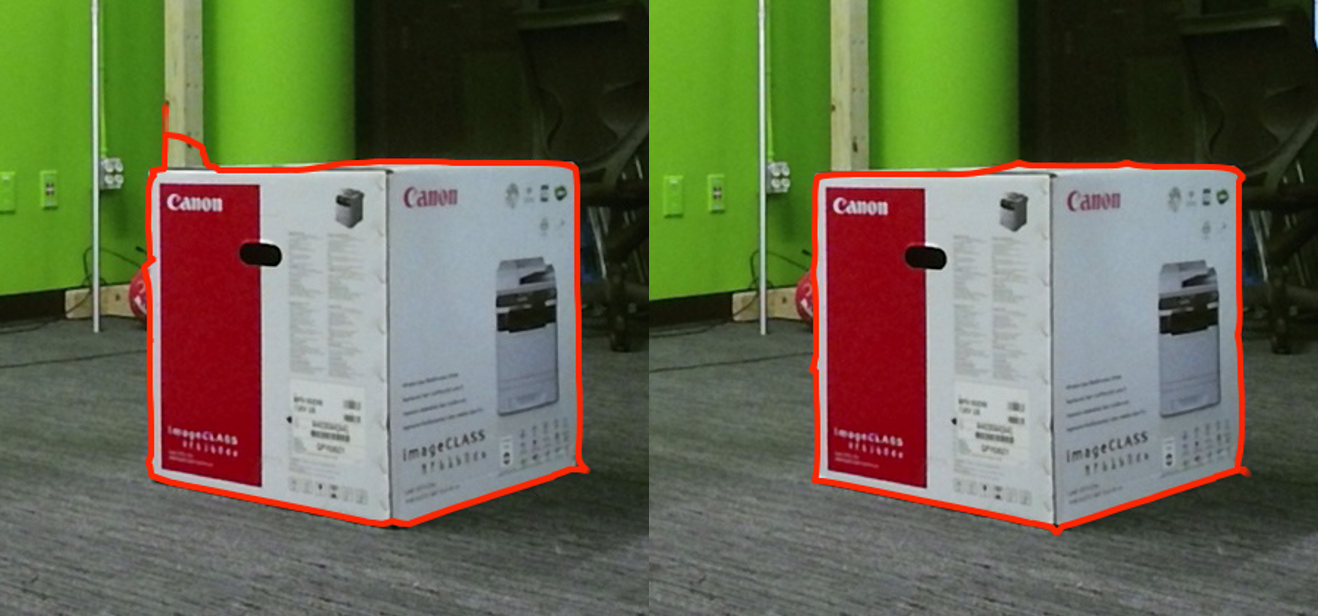}
	\caption{ Interactive region growing 2D segmentation method. Left is only using RGB image, right is using both RGB image and depth image, which keep the better boundary for the object. }
	\label{fig:segexp}
	\vspace{-1em}
\end{figure}

In the first phase of our method, firstly user uses the haptic avatar to touch and mark an object of interest in a 3D scene. Then the 3D mark is transformed to the current color image coordinates. At the next time step, starting from the mark point in the image, the pixels are clustered through a region growing method until there are no pixels to be added. The region is treated as a cluster, then the distance between the neighboring pixels and the cluster center is computed as the combination of two Euclidean distances as shown in the Equation~\ref{eq:segdistance}:
\begin{equation}\label{eq:segdistance}
d(\mathbf{x_i}, \mathbf{S}) = \|I(\mathbf{x_i}) - I(\mathbf{S})\|_2 + \beta\|P(\mathbf{x_i}) - P(\mathbf{S})\|_2
\end{equation} 
where $\mathbf{x_i}$ is the neighbor pixel position and $\mathbf{S}$ is the center of the region. $I$ is the CIELAB color space value of the pixel in the color image, which is widely considered as perceptually uniform for small color distances~\cite{achanta2010slic}. $P$ are the 3D coordinates that are computed from the depth image. The values for cluster center: $I(\mathbf{S})$ and $P(\mathbf{S})$ are computed as the averages of the values of all pixels in this cluster. $\beta = \frac{m}{g}$ is a parameter that controls the compactness of a region. $m$ is the variable to control the compactness, $g$ is the grid interval.
we first carried out an experiment comparing the region growing with RGBD data and only with RGB data, as shown in Figure~\ref{fig:segexp}. With RGBD data, the boundary of the object is kept better than that only using RGB data.

The greater the value of $m$, the more spatial proximity is emphasized and the more compact the cluster. 
This value can be in the range [1, 20]. We choose $m = 10$ for all the results in this paper. The distance threshold can be chosen by the user.  

\section{Haptic-enabled Mixed-initiative Control}
\label{sec:hapticcontrol}

In most previous works~\cite{sauer2011mixed, cacace2014mixed}, haptic force feedback is used to generate path guidance forces, collision preventing forces, or environmental force to improve the performance of high-level tasks operations. However, our system uses a haptic device in a different way. The haptic device is used as the 3D avatar to remotely ``touch", explore and interact with the virtualized real-world environment. The haptic interaction provides more flexible operations similar to using "virtual hands". 

\subsection{Haptic Interfaces}
The haptic interfaces can intervene in the robot control procedure, and add a new path or change destinations. These interfaces will not influence the velocity, but only the paths and target points.

\begin{figure}[htb!]
	\centering
	\includegraphics[width = 0.7\linewidth]{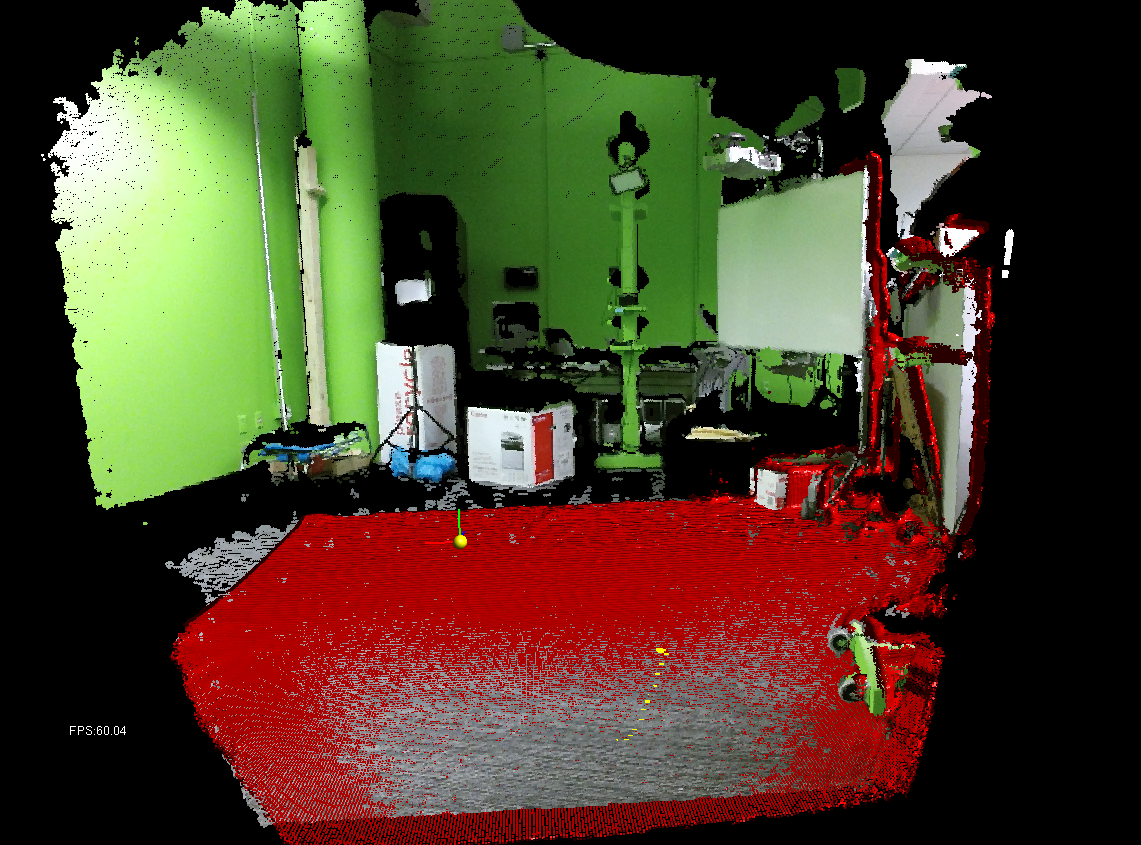}
	\caption{Haptic marking interface, the user can touch the floor, and mark a path (yellow color) on the point cloud.  }
	\label{fig:marking}
	\vspace{-0.5em}
\end{figure}

\noindent\textbf{Haptic Marking for Path Guidance}
Since haptic rendering with the surface is in real-time and efficient, our system provides a haptic marking interface. The user can use HIP to touch the floor to mark a path. Then the control manager takes this marked path as input to invoke path planning. The marking is saved as the ordering point sets and saved separately in the remote server. The interface is shown as Figure~\ref{fig:marking}. 

\begin{figure}[htb!]
	\centering
	\includegraphics[width = 0.7\linewidth]{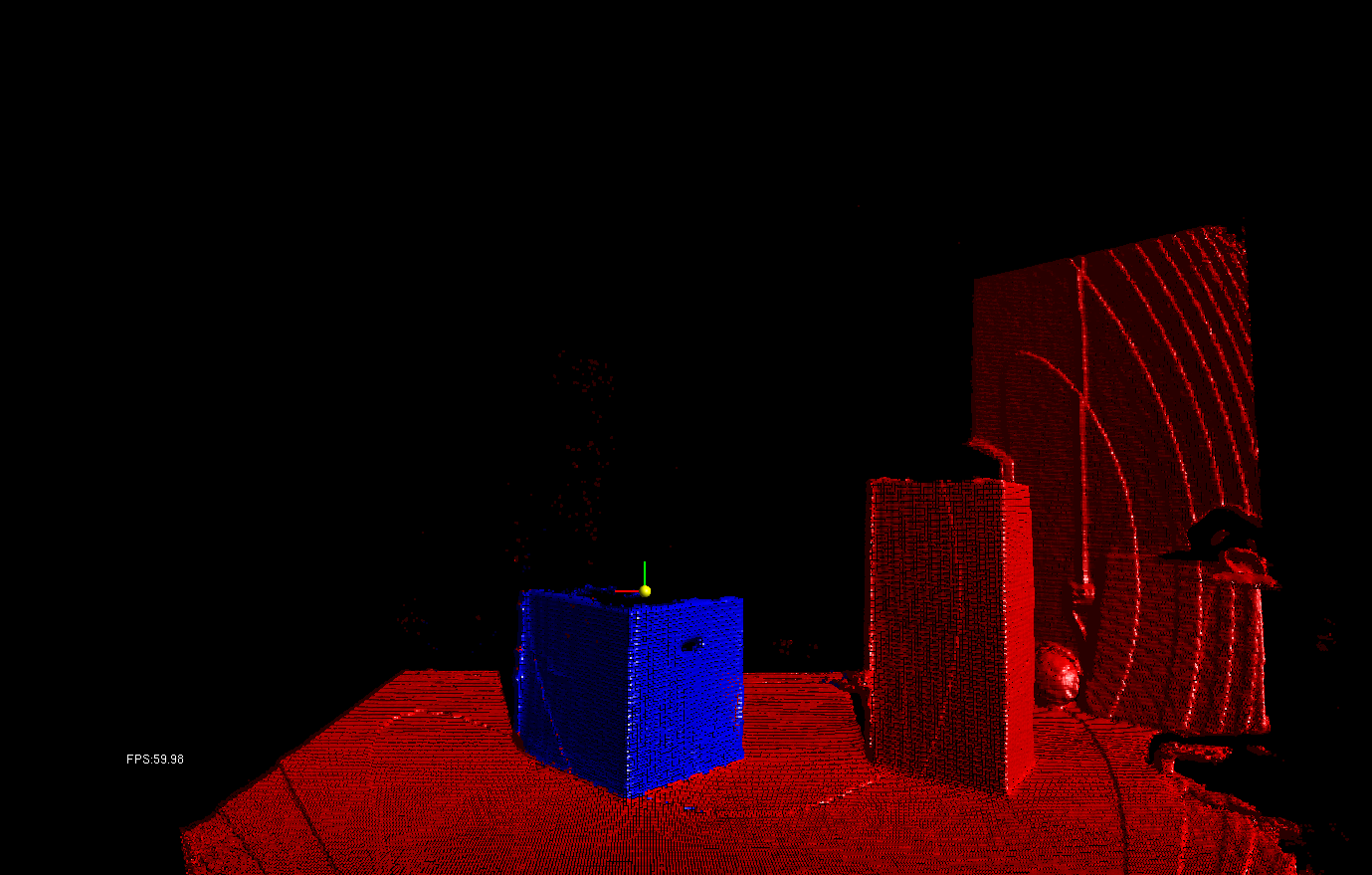}
	\caption{After object segmentation, set this object to be target (blue).  }
	\label{fig:segmark}
	\vspace{-1em}
\end{figure}

\noindent\textbf{Haptic Marking for Object}
When the user wants the robot to approach an object and grab it, the user can use the interface to first segment the object, and then set the object as the target to approach. The snapshot of this marking is shown in Figure~\ref{fig:segmark}.  

\begin{figure}[htb!]
	\centering
	\includegraphics[width = 0.7\linewidth]{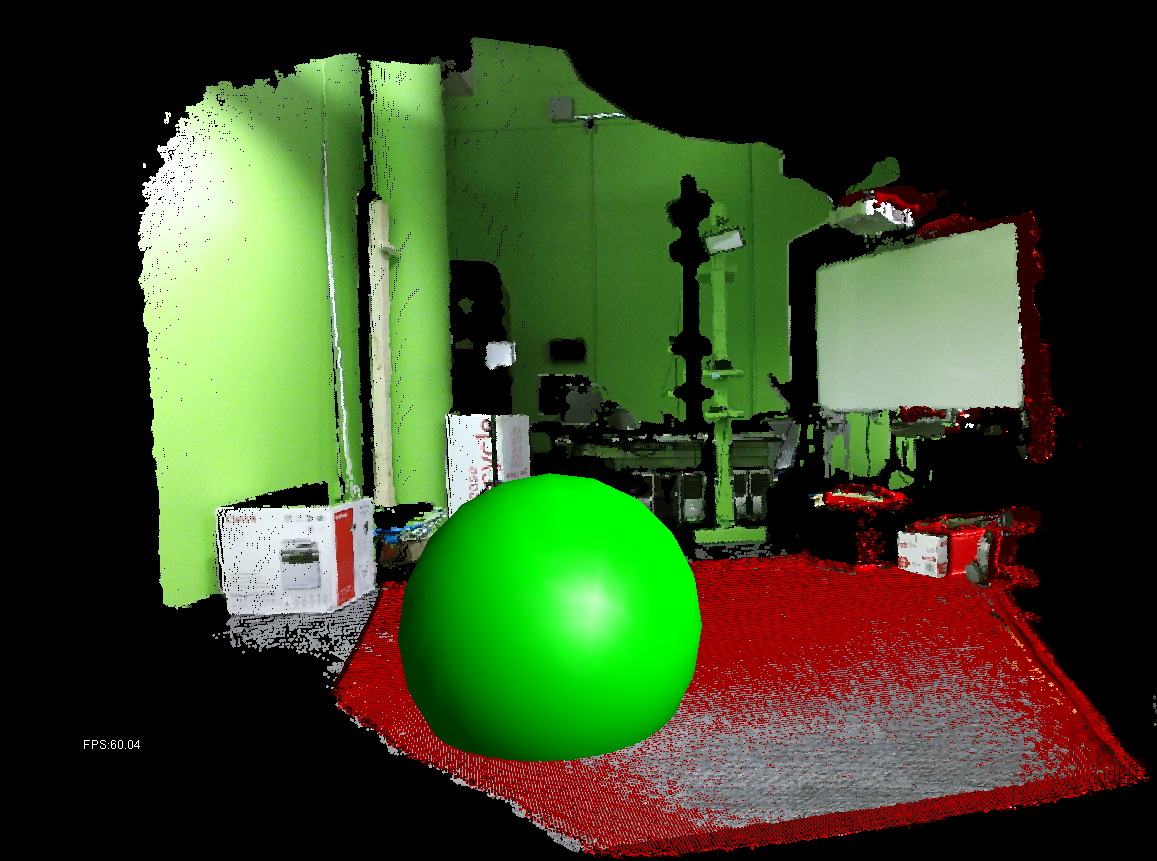}
	\caption{User use the haptic cursor to place a virtual obstacle (green sphere). }
	\label{fig:obstacle}
	\vspace{-1em}
\end{figure}

\noindent\textbf{Virtual Obstacle}
Some researchers have used augmented reality to set up virtual objects~\cite{sauer2011mixed}. In our system, this operation is much easier since a haptic device can locate a 3D position fast and accurately. The proposed system provides an interface that users can put virtual obstacles on the ground. The ground plane is located and saved at the first several time steps. The virtual objects can be treated as obstacles, and the path planner will regenerate the new path to avoid them. The snapshot of this obstacle is shown in Figure~\ref{fig:obstacle}.

\begin{figure}[htb!]
	\centering
	\includegraphics[width = 0.8\linewidth]{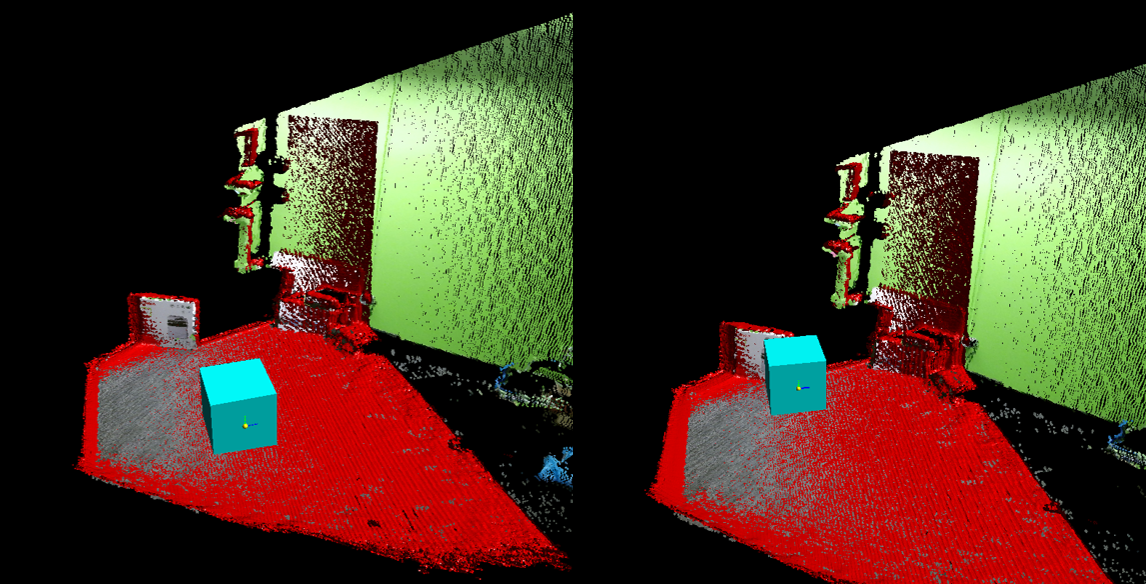}
	\caption{User use the haptic cursor to push a virtual cube. }
	\label{fig:interaction}
	\vspace{-1em}
\end{figure}

\noindent\textbf{Haptic Enabled with Physics Simulation}
In computer graphics applications, the haptic-enabled physics simulation is very popular. Since virtual objects can be placed, users can use a haptic avatar to interact with the virtual objects. These objects can be treated as new visual cues, marks, or obstacles. The interface is shown as Figure~\ref{fig:interaction}.

\subsection{Control and Latency Compensation}
Our system requires a control system with the following features: 
\begin{enumerate}[leftmargin=*, label=(\roman*)]
	\item High-level mix-initiative control needs to consider the network latency. 
	\item The system supports sliding autonomy. Both autonomous and human-in-the-loop control modalities need to be supported.
\end{enumerate}

\begin{figure}[htb!]
	\centering
	\includegraphics[width = 1.0\linewidth]{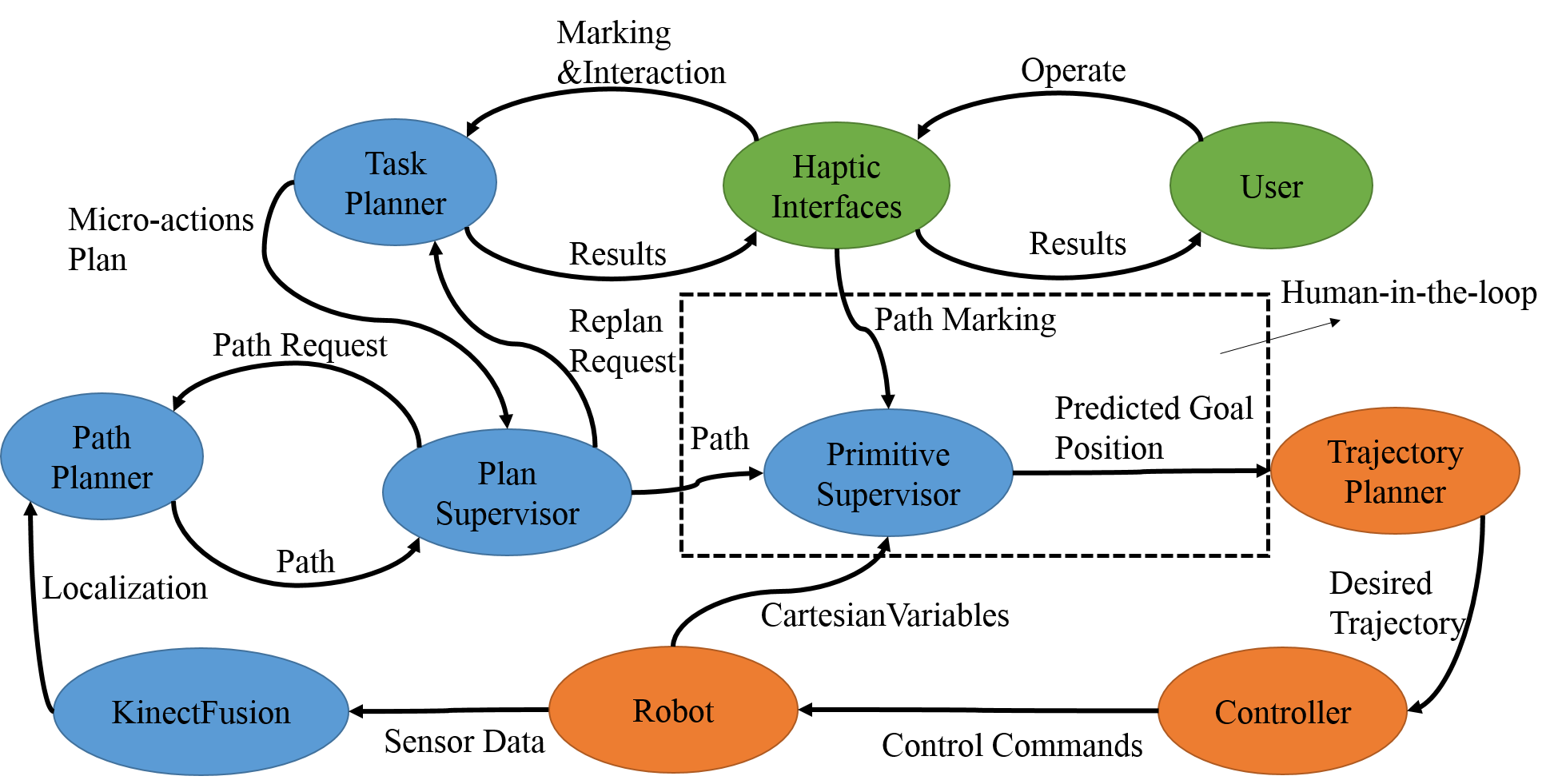}
	\caption{The proposed control architecture, Execution Level and Robot Level are listed as blue and orange respectively. }
	\label{fig:control}
	\vspace{-1em}
\end{figure}

To incorporate these features, the control architecture will be distributed in both Robot Layer and Execution Layer, which is shown in Figure~\ref{fig:control}. The Execution Layer includes Task Planner, Path Supervisor, Path Planner, and Primitive Supervisor. The Robot Layer includes Trajectory Planner, Controller, and the Robot. 
The user uses haptic interfaces to invoke high-level tasks, including haptic marking a position or an object, haptic interaction with a virtual object. These operations are passed to Task Planner. Task Planner is a high-level manager to communicate with the Plan Supervisor. It can parse the task into the micro-actions plan, and receive the replanning request. Plan Supervisor can request and receive the path between two points from Path Planner. In our framework, the path generation is based on a Rapidly-exploring Random Tree algorithm~\cite{lavalle1998rapidly}. 

In our mixed-initiative control, human-in-the-loop happens in this Primitive Supervisor module, as shown in Figure~\ref{fig:control}.
The low-level Primitive Supervisor receives the path information such as waypoints and micro-actions from the Plant Supervisor. It will receive the planned path, and also the haptic marking path, and generate a goal position for the robot motion. In the Robot Layer, Trajectory Planner monitors and controls the trajectory towards the goal position. The haptic marking path provides a marking point $x^m$, and the planned path provides a path point $x^p$. The goal position $x^g$ is chosen from these two points by choosing the maximal distance between the point and the current robot position. Network delays may influence the mapping and localization from KinectFusion. To compensate for the delay, we propose a method to generate a predicted goal position. Assuming the current velocity of the robot is $v_i$ = $(a_i, b_i, c_i)$ at $ith$ time step, the straightforward way to predict the next velocity is to compute the velocity and acceleration with the last several frames. Most Kalman filters are based on an empirical model of this linear form. In this paper, we applied the general linear model to predict the next velocity $v_{i+1}$ :

\begin{equation}\label{eq:velacc}
\begin{split}
a_{i+1} &= \alpha_0a_i + \alpha_1a_{i-1} + ... \alpha_ma_{t-m}\\
b_{i+1} &= \beta_0b_i + \beta_1b_{i-1} + ... \beta_mb_{t-m}\\
c_{i+1} &= \gamma_0c_i + \gamma_1c_{i-1} + ... \gamma_mc_{t-m}
\end{split}
\end{equation} 

For a given time series of points in a path, the matrix $V$ is defined as Eq.~\ref{eq:matrix}: 
\begin{equation}\label{eq:matrix}
\textbf{V} = 
\begin{bmatrix}
a_0 & \dots  & a_m &  b_0 & \dots  & b_m & c_0 & \dots  & c_m\\
a_1 & \dots  & a_{m+1} &  b_1 & \dots  & b_{m+1} & c_1 & \dots  & c_{m+1}\\
\vdots \\
a_i & \dots  & a_{i+m} &  b_i & \dots  & b_{i+m} & c_i & \dots  & c_{i+m}\\
\end{bmatrix}
\end{equation}
Let $\textbf{v}$ to be the predicted positions $(v_{m+1}, v_{m+2}, ..., v_{i+m+1}, ...)^T$. The problem now is to solve and obtain three parameter vectors $\alpha$, $\beta$, and $\gamma$. The general solution of these linear problem are shown as follows: 

\begin{equation}\label{eq:prediction}
\begin{split}
\alpha &= (\textbf{V}^{T}\textbf{V})\textbf{V}^{T}a\\
\beta &= (\textbf{V}^T\textbf{V})\textbf{V}^Tb\\
\gamma &= (\textbf{V}^T\textbf{V})\textbf{V}^Tc
\end{split}
\end{equation}
Every time step, this linear prediction model generates new parameters, and then predicts the next goal position $x^g_{i+1} = x^g_{i} + v_{i+1}t$, where $t$ is the round time delay. This goal position will be sent to Trajectory Planner for the low-level autonomous control.


\begin{table*}[htb!]
	\centering
	\caption{Processing time of different components using different TSDF resolution.} \label{tab:performance_streaming}
	\begin{adjustbox}{width=1\textwidth}
		\begin{tabular}{|c|c|c|c|}
			\hline
			& \multicolumn{3}{c|}{Average Processing Time}\\
			\cline{2-4}
			\raisebox{2.4ex}{TSDF Resolution} & \raisebox{1.3ex}{Total Time} & \raisebox{1.3ex}{Segmentation and KinectFusion} & \raisebox{1.3ex}{Collision Detection and Handling}
			\\
			\hline
			64*64*64 & $16.4ms$ & $11.8ms$ & $0.6ms$ 
			\\
			128*128*128 & $20.3ms$ & $14.3ms$ & $1.3ms$  
			\\
			256*256*256 & $22.2ms$ & $15.5ms$ & $1.6ms$  
			\\
			384*384*384 & $31.9ms$ & $22.5ms$ & $2.4ms$  
			\\
			512*512*512 & $38.8ms$ & $27.3ms$ & $2.7ms$  
			\\
			\hline
		\end{tabular}
	\end{adjustbox}
\end{table*}
\vspace{-1em}

\section{Haptic-enabled Mixed-initiative Control Experiment}
Here, we demonstrate the experimental setup and results. 
The remote site is an Intel i7 3.50 GHz machine with 32 GB RAM. Equipped GPU is GeForce GTX 670. For KinectFusion, the dimension of the TSDF volume is usually set to be 512*512*512, voxel size is about 10 mm, the truncated depth distance is set to be 50 mm. In KinectFusion, a 3-level iterative closest point (ICP) structure is applied by iterations 4/5/15. The local site includes a KUKA youBot mobile robot, and a Kinect V2 is placed on top of the robot. The robot can move in 4 directions, and turn by 45 degrees every time. The Kinect sensor is connected to a Linux machine with ROS for robot control. With a network operating at 10 Mbps, TCP/IP protocol is employed for the data transfer. The system transfers the RGBD images to the remote site at a rate of 15-20 fps. The testbed setup is shown in Figure~\ref{fig:hapticrobot}. We used KUKA youBot mobile robot and Force Dimension Omega.3 haptic device. Omega.3 has 3 degrees of freedom, 14.5 N/mm stiffness, and up to 12.0 N force.

\subsection{Haptic Rendering}
The TSDF resolution determines the 3D geometry density, and it will further influence the time efficiency of the proposed haptic rendering method. We carried out the experiments for haptic rendering using different TSDF resolutions. The average processing time is recorded and shown as Table.~\ref{tab:performance_streaming}. Proxy update belongs to the ``collision detection and handling". As shown in the results, all modules of our haptic rendering method are efficient. The proposed haptic rendering with KinectFusion can support the real-time mixed reality applications, even using a high TSDF resolution. 

\begin{figure}[htb!]
	\centering
	\includegraphics[width = 1.0\linewidth]{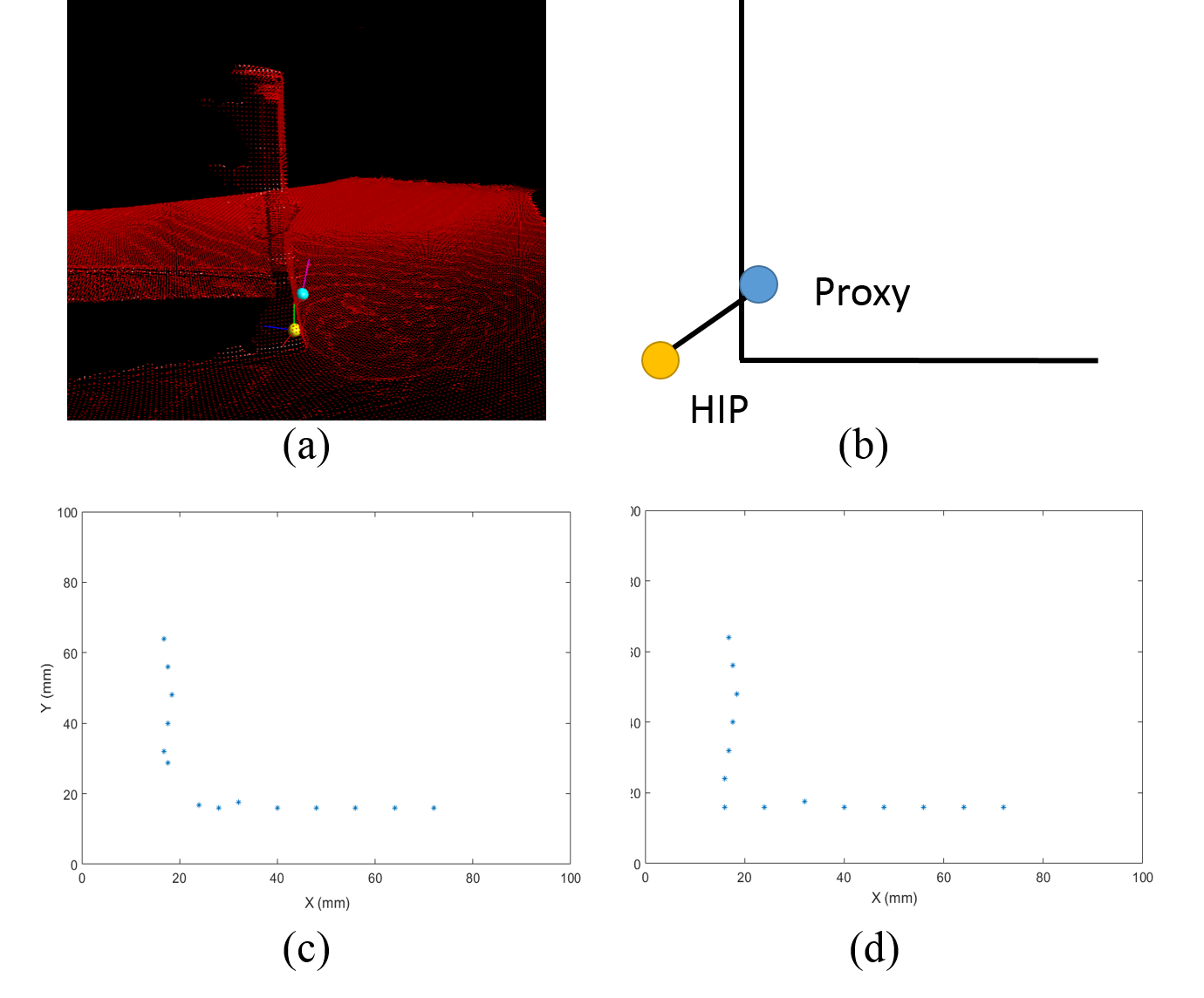}
	\caption{The accuracy of haptic rendering. (a) The side view of a boundary between front face of the box and ground. (b) Sketch of the haptic interaction with a sharp corner of two planes. (c) A two dimensional crossing section of the proxy (asteroid) positions when moving over the surface, using haptic rendering method in \cite{tian2017real}. (d) Crossing section of proxy positions using our method. }
	\label{fig:accuracy}
	\vspace{-1em}
\end{figure}

The accuracy of the haptic rendering is evaluated by using the HIP to move with a sharp boundary. In this experiment, a box is placed in the scene. The front of the box and the floor generates a sharp concave boundary. Figure~\ref{fig:accuracy} shows the experimental scenario and results. As shown in Figure~\ref{fig:accuracy}(c), haptic rendering in \cite{tian2017real} cannot lead to smooth proxy update when moving over the corner. Since the force feedback is computed based on the distance between proxy and HIP, the abrupt change of proxy leads to abrupt force change or even vibration force. Our proxy update method with force shading solves this problem and provides smooth force feedback.

\begin{table*}[htb!]
	\centering
	\caption{Comparison results of segmentation on BSDS500 benchmark.} 
	\label{tab:segmentation}
	\begin{adjustbox}{width=1\textwidth}
		\begin{tabular}{|c|c|c|c|}
			\hline
			& \multicolumn{3}{c|}{Measurement}\\
			\cline{2-4}
			\raisebox{2.4ex}{ Methods  }
			& \raisebox{1.3ex}{ Probabilistic Rand Index}
			& \raisebox{1.3ex}{Boundary Displacement Error} & \raisebox{1.3ex}{ Global Consistency Error}
			\\
			\hline
			Normlized Cut & $0.73931$ & $17.1560$ & $0.2232$ 
			\\
			CTM & $0.7796$ & $19.1981$ & $0.3647$ 
			\\
			Meanshift & $0.7769$ & $13.1616$ & $0.5811$  
			\\
			Our method & $0.73876$ & $14.216$ & $0.5432$   
			\\
			\hline
		\end{tabular}
	\end{adjustbox}
\end{table*}

\subsection{Interactive Segmentation}
In our system, users are free to interact with the whole surface or one specific object surface. 
To evaluate the interactive object segmentation method, we compare the 2D image segmentation on the BSDS500 benchmark~\cite{arbelaez2011contour}. This database consists of natural
images with five different human ground truth segmentations. In the comparison, we use the mouse to mark the starting pixel instead of the haptic device. We compare the results with popular segmentation methods: normalized cut~\cite{shi2000normalized}, meanshift~\cite{comaniciu2002mean} and Compression-based Texture Merging (CTM)~\cite{yang2008unsupervised}. The comparison result is shown in Table.~\ref{tab:segmentation}. The comparison results show our method has comparable performance with the other popular segmentation methods. We use 3 performance measures: Probabilistic Rand Index (the fraction of pairs of pixels whose labels are consistent between the computed segmentation and the ground truth), Boundary Displacement Error (the average displacement error of boundary pixels between two segmented images) and Global Consistency Error (the extent to which one segmentation can be viewed as a refinement of the other)~\cite{deng2001unsupervised}.  

\subsection{Control and Execution}
\begin{figure}[htb!]
	\centering
	\includegraphics[width = 0.8\linewidth]{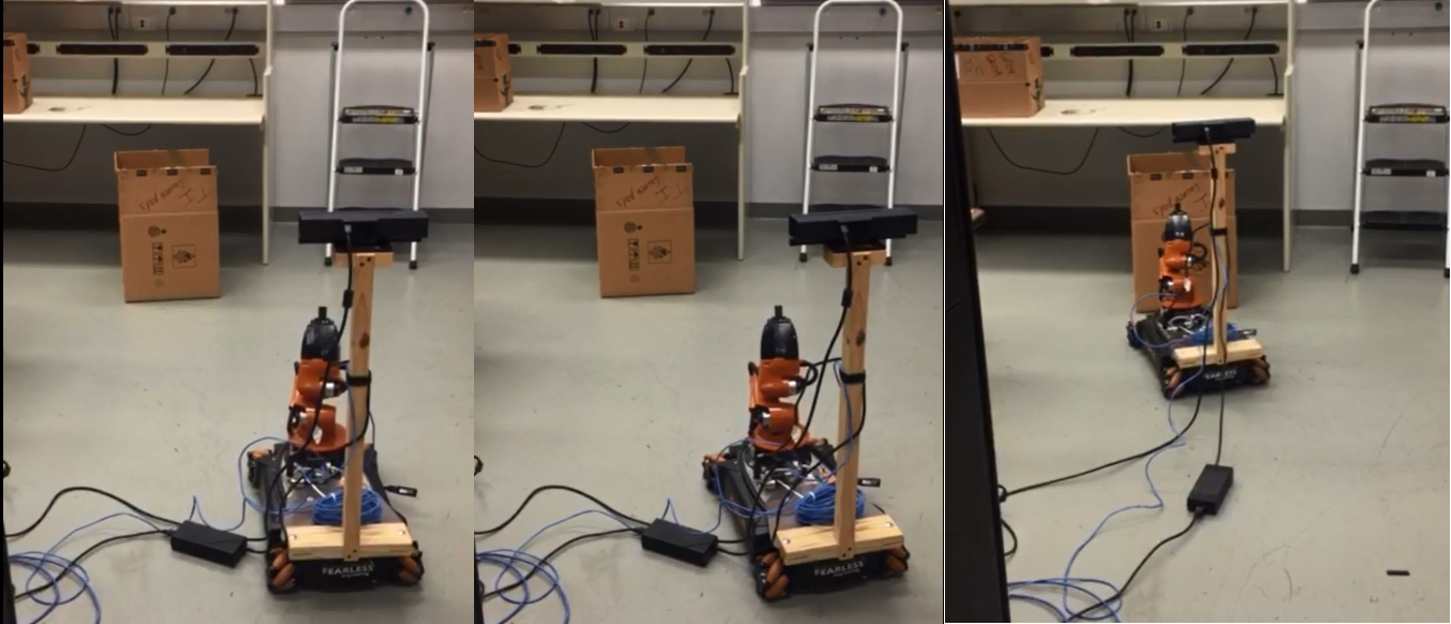}
	\caption{The haptic marking control scenario. After the user set target to be the box, the robot approach the box. }
	\label{fig:robotcontrol}
	\vspace{-1em}
\end{figure}

\begin{figure}[htb!]
	\centering
	\includegraphics[width = 0.9\linewidth]{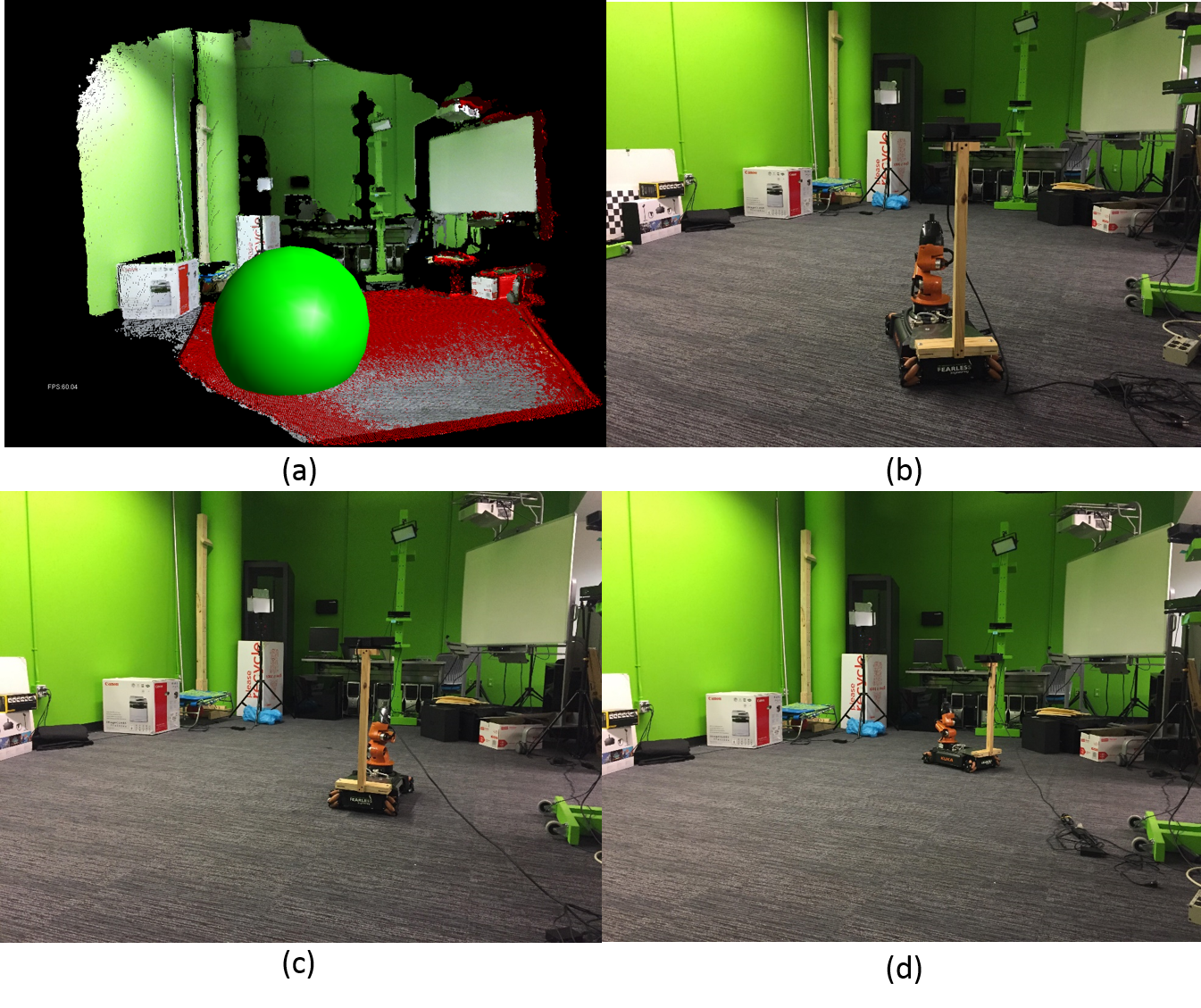}
	\caption{ (a) The user firstly defines a target, then adds a virtual obstacle (green ball) into the scene. (b)(c)(d) The robot find the path to avoid the obstacle. }
	\label{fig:obstacle_result}
	\vspace{-1em}
\end{figure}

We have carried out an experiment to testify the system control and execution performance. The robot control loop is performed with a small delay (less than 60 ms). The first control scenario is to approach the target after it is marked, as shown in Figure~\ref{fig:robotcontrol}. The second scenario is to add the virtual obstacle in the scene, as shown in Figure~\ref{fig:obstacle_result}. We involved 2 students and each subject was asked to repeat the experiment 5 times
in the testbed. For both scenarios, we executed 10 times.  We have collected the mean, min, max, standard deviation (STD) of time of planning (Tp), time of replanning (Tr), length of the executed path (Lp), and a total time of execution (Te). The results show that the control/execution performance is compatible with the operative scenario requirements~\cite{wen2012robot}. Since we focus on haptic-enabled mixed-mediated control, the control optimization is beyond our scope. 

\subsection{Latency Compensation}
\begin{figure}[htb!]
	\vspace{-1em}
	\centering
	\includegraphics[width = 0.8\linewidth]{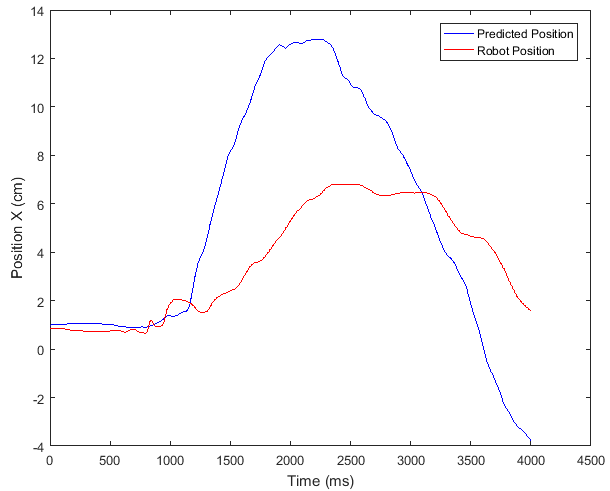}
	\caption{The comparison between the real-world robot position curve and predicted position curve with 200ms network latency. }
	\label{fig:curve}
	\vspace{-1em}
\end{figure}

\begin{figure}[htb!]
	\centering
	\includegraphics[width = 0.8\linewidth]{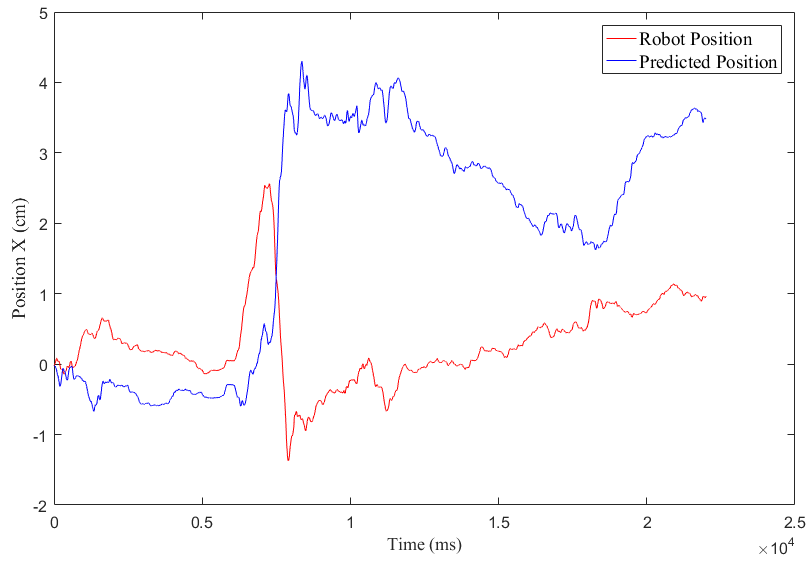}
	\caption{The comparison between the real-world robot position curve and predicted position curve with time varying network latency from 100ms to 200ms. }
	\label{fig:curve2}
	\vspace{-1em}
\end{figure}

To verify the delay compensation method, we applied two delays (100ms, 200ms) over the Internet. We recorded the real-time robot positions and also the predicted positions generated from the proposed linear model.
The error is defined as the Euclidean distance along X-axis since robots are moving along the x-axis in two experiments.
If big error means the newly planned position sent from server to robot is very far from the real position. Based on the control test, if this error is beyond 20cm then the trajectory will be influenced. 
As shown in Figure~\ref{fig:control}, the trajectory planner might generate a costly trajectory, and the robot may move back and forth because of the delay.
The robot positions are computed for each frame from the odometry. Figure~\ref{fig:curve} shows two-position curves (along X-axis) over the Internet with 200ms latency. 
The average position error for 100ms is 2.34cm, and 3.47cm for 200ms. The maximal error for both latencies is lower than 7.0cm, which shows good control performance.
To further test the latency, we have one more experiment that adding time-varying network delay from 100ms to 200ms as shown in Figure~\ref{fig:curve2}. The difference between the two curves is very close, and lower than 7.0cm. 
These results show that our latency compensation can handle up to 200ms network latency. 
\begin{table}[htb!]
	\centering
	\caption{Control and Execution Results.} 
	\label{tab:control}
	\begin{adjustbox}{width=0.4\textwidth}
		\begin{tabular}{|c|c|c|c|c|}
			\hline
			& \multicolumn{4}{c|}{Control Measurements}\\
			\cline{2-5}
			\raisebox{2.4ex}{ Method}
			& \raisebox{1.3ex}{ Mean}
			& \raisebox{1.3ex}{ STD}
			& \raisebox{1.3ex}{Max} & \raisebox{1.3ex}{Min}
			\\
			\hline
			Tp & $0.082s$ & $0.013s$ & $0.09s$ & $0.04s$
			\\
			Tr & $0.614s$ & $0.43s$ & $1.70s$ & $0.01s$ 
			\\
			Te & $70.8s$ & $25.1s$ & $83s$ & $42s$  
			\\
			Lp & $15.4m$ & $2.35m$ & $19m$ & $12m$  
			\\
			\hline
		\end{tabular}
	\end{adjustbox}
\end{table}
\vspace{-1em}

\section{Mixed-Initiative Control for dexterous robotic hand for various applications -a perspective}

The Kuka YouBot robot used in the previous section is great for applications requiring maneuvering in-house and precise control. However, the robotic arm has a limited degree of freedom and will not be able to replicate human-like actions from remote locations. Here, we describe the essential elements that are needed for improving the capabilities of our current work and use them to solve engineering/healthcare problems. These are listed as follows: 1) designing a robotic hand with five fingers, 2) Use the hands for special operations such as healthcare and military applications. The general healthcare and military applications have their own requirements and hence require unique solutions.

\subsection{Dexterous Humanoid Hand Design and Control for Complex Tasks in Healthcare and Hazard Mitigation}

Five-fingered robotic hands are needed in many applications to imitate human action in a remote location. We have designed such hands using the 3D printing method and using parametric-based CAD software~\cite{wu2017compact,lanigan2017low}. The parametric-based hand design can be easily customized and scaled as needed, we can print big adult size hands or small child size. It is a low-cost but highly dexterous robotic hand that can carry out complex tasks, the robotic hand can be actuated using novel actuators such as TCP muscles. Twisted and Coiled Polymer (TCP) muscles or actuators are soft polymers that enable the realization of low-cost and high-performance humanoid robots~\cite{wu2017compact,wu2015nylon}. TCP muscles contract when heated and return to their original shape, like natural muscles. They do not produce any noise, as a result, can be used for silent operations. This is the key advantage of designing robotic arms using soft actuators rather than motors and pneumatics.  We have been studying the TCP materials to develop a novel musculoskeletal system. TCP muscles could provide large strain (20-49\%), large stress (1-35MPa) and high mechanical work (5.3 kW/kg)~\cite{wu2017compact,haines2014artificial}. More importantly, the material cost for making the muscle is low compared to shape memory alloy actuators~\cite{haines2014artificial}. Therefore, it is worth studying this material further to develop high-performance and low-cost robots including the closed-loop control systems~\cite{jafarzadeh2018control}.

\begin{figure}[htb!]
	\vspace{-1em}
	\centering
	\includegraphics[width = 0.8\linewidth]{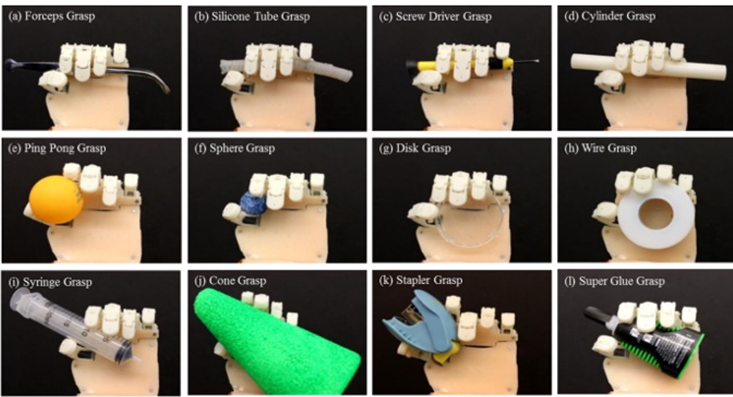}
	\caption{Grasping objects using  our robotic hand, UTD Hand that is made out  of TCP muscles~\cite{wu2017compact}. }
	\label{fig:grasping}
	\vspace{-1em}
\end{figure}

When referring to literature, several robots and robotic hands have been developed in universities and research institutes~\cite{diftler2011robonaut,hanson2006enhancement,kajita2011cybernetic,oh2006design}.
The actuators typically used in these robots are expensive and are not biomimetic. Some of the advanced humanoids include Boston Dynamics’s Atlas, ASIMO, Robonaut, HUBO and HRP-4C. These robots are extremely expensive and most of them are not available commercially. For example, ASIMO costs about \$1 million and even \$100,000 for a rent~\cite{sofge_2018} mimetic and affordable. We have made several efforts in the last 10 years in creating humanoids using various smart actuation technologies: piezoelectric, conducting polymer, shape memory alloy actuators and the latest humanoid hand that is actuated by TCP muscles and using additive manufacturing technology Figure~\ref{fig:grasping}. The key feature of the design is that 1) it can grasp various daily used objects, 2) the actuators do not require large space , they are kept in the forearm, 3) The structure is lightweight as it is polymer based and 4) No electromagnetic noise generated. Key scientific challenges and performance of smart materials were described in our previous works~\cite{tadesse2013electroactive,tadesse2011twelve,tadesse2009synthesis,jafarzadeh2021wearable}. The TCP muscles are used for actuation of fingers of the robots~\cite{wu2017compact,wu2015nylon} and hence we will investigate this hand with the proposed mixed-initiative method in the future.

\begin{figure}[htb!]
	\centering
	\includegraphics[width = 0.8\linewidth]{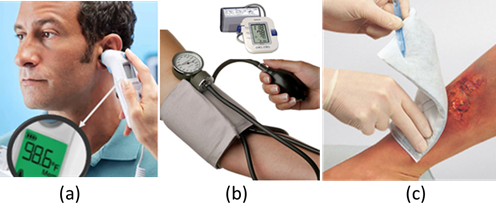}
	\caption{Robots for use in healthcare innovation: (a) temperature measurement using Braun ThermoScan\textsuperscript{®}, (b) pressure measurement of a subject, (c) wound cleaning.}
	\label{fig:healthcare}
	\vspace{-1em}
\end{figure}

\subsection{Dexterous Hand for Complex Tasks in Healthcare Application via Teleoperation}

Our main objective is to show the benefit of maturing technologies presented in this paper, identifying and solving the key challenges in healthcare. The goal is to develop a low-cost but highly dexterous robotic hand that can carry out complex tasks, especially those that might be dangerous for humans such as handling contagious diseases. This aspect makes the proposed command and control of a robotic hand transformative because of the tradeoff between cost, overall system size, weight, operation noise, and performance in handling objects. We would like to show the use of the proposed solution in three different cases that require their own requirements.

\textbf{Case 1: Demonstration of Command and Control in Healthcare- Prolonged Field Care (PFC)}

In some applications, robots are desired to monitor and help individuals in need in remote field geographic location for an extended period of time. This could be physical assistance combined with visual monitoring. Our demonstration will be focused on prolonged field care (PFC) and well suited in this area particularly for especial cases, an epidemic disease that is not safe for medics,  because our proposed robot can monitor the subject 24/7 as well as act as needed (such that help the patients, providing water, food and other items, (Figure~\ref{fig:healthcare} a-c).


We will have two main applications and research efforts: (i) body temperature measurement, and (ii) wound management in field care units that the military needs. Our solution is particularly useful for special missions that are difficult or unsafe for humans to do.   

\textbf{Case 2:  Body temperature measurement from Remote Location}

\begin{figure}[htb!]
	\vspace{-1em}
	\centering
	\includegraphics[width = 0.8\linewidth]{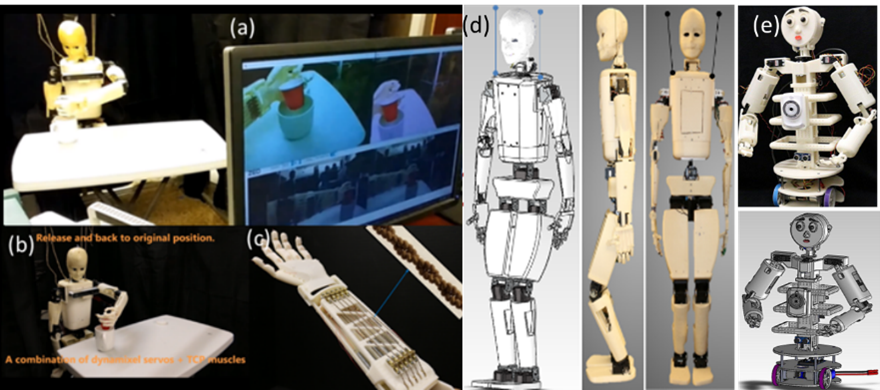}
	\caption{Robotic hand with TCP muscles, picking and placing an object: (a) The hand-powered with TCP muscle and the inset of the muscle, (b) Our 3D printed humanoid robot actuated by servomotors and TCP powered hands; (c) holding and moving an object - as seen by the 2 fire-wire cameras in the head and the 3D camera in the chest; (d) HBS 1 Biped; (e) Buddy wheeled robot. }
	\label{fig:humanoids}
	\vspace{-1em}
\end{figure}

Another great application of the proposed method in this paper is taking a body temperature measurement from a human subject via teleoperation. We would like to experiment with the hand developed in the previous section to take temperature measurement of a patient simulator using hand-held Braun ThermoScan® (Figure~\ref{fig:healthcare} a). Patient simulators are a great way to expedite such research as human subject tests typically require multiple processes of IRB approval at this initial stage. First, we will perform simple experiments to let the robot grasp ThermoScan and manipulate it by developing algorithms. Inverse kinematics~\cite{parga2013smartphone,sciavicco2012modelling,spong2008robot}, camera and hand coordination will be employed to direct the thermometer to the patient simulator’s forehead, and the robot will read the instrument display. We will use Object Character Recognition (OCR)~\cite{mori2003recognizing} program to detect the actual temperature reading.  The temperature and other procedures require multiple actions of the robot following particular algorithms and experiments. We will also determine the reachable positions (workspace) of the robot hand using the Denavit-Hartenburg (DH) method, a convention used to represent the relationships of linkage parameters in robotic manipulators.  We have done some preliminary tests on the actuation of the robot locally, without using the teleoperation, to see the practical issues and identify the actuation variable. Some of the prior works are presented in \cite{wu2018biorobotic}. Our humanoid robotic with the specialized hands (Figure~\ref{fig:humanoids} c) will be mounted on a mobile and it can accomplish such a task by performing a series of experiments.  We will customize our robots to perform these tasks.  We have some representative video of our robot can be found from https://youtu.be/WKc32gcdgj0.

\textbf{Case 3: Low-cost and High-performance Biped Robots and Mobile 3D Printed robots-for Hazardous Substance}

Low-cost humanoids are needed to manipulate hazardous substances and explosives. They can be made again if it is destroyed during operations or field trials.  Our team has designed and developed a 3D printed robot, HBS-1 robot (Figure~\ref{fig:humanoids} a) that has mechanical systems, electrical and mechatronic systems. The materials cost of this robot is \$10,000-\$15,000~\cite{wu2017compact,wu2016hbs}. The overall dimension of robot HBS-1 is 120 cm x 33cm x 14 cm, which match closely a 7-year-old boy~\cite{snyder1975anthropometry}. HBS-1 consists of two 4-DOF legs, a 2-DOF waist, two 4-DOF arms, two 15-DOF hands, and a 3-DOF head (51 DOF in total). The robot is powered by a DC power supply connected through wires while tethered. HBS-1 utilizes 14 Dyanmixel servos. Shape memory alloy and TCP actuators have been used for the design of the fingers since they can be installed in the limited volume of the forearm. Otherwise, it would have been difficult to actuate all five fingers. HBS-1 is equipped with Firewire stereo cameras which are housed in the head. The torso is equipped with an orientation sensor (UM7-LT), which combines gyroscopes, accelerometers, magnetic sensors, and an onboard 32-bit ARM Cortex processor to compute sensor orientation. Two servo controllers in the torso actuate the 21 servo motors.
The other robot called Buddy (Figure~\ref{fig:humanoids}b) consists of a wheeled mobile base, a cloud camera, an ultrasonic position sensor, battery, wireless communication module, flexible touch sensor skin, two 4-DOF arms, and a 2-DOF neck (15-DOF in total). The overall dimensions of the robot are a height of 580 mm, an arm space of 925 mm, a shoulder width of 230 mm, and a chest thickness of 172.5 mm~\cite{burns2014mechanical,potnuru20163d}. These components took 56\% of the total material cost. The total material cost of this robot is \$3000 including the mechanical and off-the-shelf mechatronic components. This robot can be modified and used for experimenting the teleoperation and handling dangerous substances.

Overall, 3D printability, dexterity, mobility are some of the key components of low-cost, high-performance teleoperated robotic systems in the three case studies presented.  The other important aspect is the 3D printed robotic hand that is actuated by coiled and twisted polymer muscles. We have recently reported such an innovative robotic hand in 2017~\cite{wu2017compact,wu2018biorobotic,karami2020modeling,saharan2020modeling}. The hand called UTD Hand can grasp various objects, which was made of inexpensive polymer muscles (Twisted and Coiled Polymer) TCP muscles. The muscles are reported in Science Magazine in~\cite{haines2014artificial}. The robotic hand can handle and manipulate various objects of different sizes and shapes which will guarantee the success of this project as shown in~\cite{wu2017compact}.  This robotic hand was featured as a unique design in the review of robotic hands in the last century~\cite{piazza2019century}. We will use this robotic hand combined with mixed-initiative teleoperation to achieve better performance.

\vspace{-1em}
\section{Conclusion and Discussion}
In this paper, we have proposed a haptic-enabled mixed reality system for mixed-initiative remote control. The system provides an efficient haptic rendering method with smooth and stable force feedback. The haptic rendering also supports the simulation of the surface properties such as friction and texture. The haptic rendering enhances the immersive environments and supports more haptic user interfaces for more flexible control commands such as pushing a virtual obstacle to change the robot's motion. 
An interactive 3D object segmentation method is also provided to segment objects fast and accurately. This segmentation result can be treated as the input for the high-level semantic classification of objects. The system also provides different types of haptic interactions in the mixed reality platform, and a prediction method to compensate for network delays. The delay can be compensated to support the complex interaction such as placing obstacles during the robot movement. The experimental results show the efficiency and functionality of our system. This system expands the user interfaces using haptic devices.  
In the future, the following user study will be performed to compare different user interface configurations. It is very meaningful to verify which haptic interface mode contributes more to task completion. Also, the prediction method can be tuned by using more complicated methods such as a non-linear model. Haptic-guided segmentation can be augmented by incorporating a learning method for semantic labeling. Our system provides haptic-guided mixed-initiative control. It can be naturally extended to more levels of autonomy using our mixed reality platform.


%

%

%


\vspace{-1em}

\bibliographystyle{IEEEtran}
\bibliography{reference,robot,segmentation,humanoid,ieee}

\end{document}